\documentclass[conference]{IEEEtran}
\IEEEoverridecommandlockouts
\usepackage{cite}
\usepackage{amsmath,amssymb,amsfonts}
\usepackage{algorithmic}
\usepackage{graphicx}
\usepackage{textcomp}
\usepackage{xcolor}
\def\BibTeX{{\rm B\kern-.05em{\sc i\kern-.025em b}\kern-.08em
    T\kern-.1667em\lower.7ex\hbox{E}\kern-.125emX}}
\begin{document}

\title{Towards advanced robotic manipulation
\thanks{This publication has emanated from research supported by Science Foundation Ireland (SFI) under Grant Number SFI/12/RC/2289\_P2, co-funded by the European Regional Development Fund.}}

\author{\IEEEauthorblockN{Francisco Roldan Sanchez}
\IEEEauthorblockA{\textit{Dublin City University} \\
\textit{Insight SFI Research Centre for Data Analytics}\\
Dublin, Ireland \\
francisco.sanchez@insight-centre.org}
\and
\IEEEauthorblockN{Given Name Surname}
\IEEEauthorblockA{\textit{dept. name of organization (of Aff.)} \\
\textit{name of organization (of Aff.)}\\
City, Country \\
email address or ORCID}
\and
\IEEEauthorblockN{ Given Name Surname}
\IEEEauthorblockA{\textit{dept. name of organization (of Aff.)} \\
\textit{name of organization (of Aff.)}\\
City, Country \\
email address or ORCID}
\and
\IEEEauthorblockN{Given Name Surname}
\IEEEauthorblockA{\textit{dept. name of organization (of Aff.)} \\
\textit{name of organization (of Aff.)}\\
City, Country \\
email addr
ess or ORCID}
}
\author{\IEEEauthorblockN{Francisco Roldan Sanchez\IEEEauthorrefmark{1}\IEEEauthorrefmark{2}\IEEEauthorrefmark{4}, \\
Supervised by: Stephen Redmond\IEEEauthorrefmark{1}\IEEEauthorrefmark{3}, Kevin McGuinness\IEEEauthorrefmark{1}\IEEEauthorrefmark{2} and Noel O'Connor \IEEEauthorrefmark{1}\IEEEauthorrefmark{2} }
\IEEEauthorblockA{\IEEEauthorrefmark{1}Insight SFI Research Centre for Data Analytics\\}
\IEEEauthorblockA{\IEEEauthorrefmark{2}Dublin City University}
\IEEEauthorblockA{\IEEEauthorrefmark{3}University College Dublin}
\IEEEauthorblockA{\IEEEauthorrefmark{4} Email: francisco.sanchez@insight-centre.org}}

\maketitle

\begin{abstract}
Robotic manipulation and control has increased in importance in recent years. However, state of the art techniques still have limitations when required to operate in real world applications. This paper explores Hindsight Experience Replay both in simulated and real environments, highlighting its weaknesses and proposing reinforcement-learning based alternatives based on reward and goal shaping. Additionally, several research questions are identified along with potential research directions that could be explored to tackle those questions.
\end{abstract}

\begin{IEEEkeywords}
robotic manipulation, deep reinforcement learning, artificial intelligence
\end{IEEEkeywords}

\section{Introduction}

Robotic control, planning, and manipulation have been attracting increasing attention in recent decades \cite{review1,rl_review,ml_review}, but many problems remain to be addressed. Since object manipulation is a trivial task for humans, we expect robots to demonstrate  comparable proficiency. However, while robots can excel in manipulating objects in a repetitive setup (e.g., factory-like conditions), where they can exploit known world priors (geometry, weight, position, etc.), they struggle in more challenging unstructured environments \cite{challenges}.

The complexity of robotic manipulation requires integrating interdisciplinary knowledge in order to reproduce a wide variety of human manipulation skills. Therefore, it is important to understand that manipulation is multimodal for humans. Vision, and even audition, is employed in the pre-grasping phase, or the \textit{planning} phase; while other sensory inputs are used during the \textit{manipulation} phase (using mainly tactile, but also visual and auditory information) \cite{challenges}. 

Traditionally, robotic manipulation tasks have been solved by dividing the problem into \textit{subtasks} and solving each subtask independently \cite{grasp_planning, traditional_example, rrc2020}. These techniques rely mainly on solving inverse kinematic equations \cite{kinematics}; i.e., finding what input parameters will move the gripper of a robotic arm along a trajectory to a desired target position and/or orientation. However, this problem has multiple solutions if the trajectory is not fully specified. This is why \textit{motion primitives} \cite{motion_prim} are typically used. These are a set of pre-computed robotic motions that have to be manually designed by an engineer. By sequentially employing these primitives, the robot can perform complex tasks. 

However, recent advances in deep reinforcement learning (RL) have proved to be promising for robotic control \cite{ddpg,dher,her}. The goal of these frameworks is to train a policy network that predicts the sequence of actions a robot must perform to successfully complete a certain task. In this approach, an agent (robot) explores the environment and receives feedback in the form of rewards, in a similar way that dogs are trained. These rewards can be dense (e.g., Euclidean distance to the target position) \cite{dense_reward_manipulaton} or sparse (e.g., 1 if the task is completed, 0 otherwise). Even though dense rewards are generally preferred because they carry more information, these are hard to define, particularly for manipulation tasks, reason why sparse rewards are widely used in this context. 

In this paper, the most popular state of the art technique for robotic manipulation is explored in order to identify its potential weaknesses, as well as the modifications  required when it is applied on a real robot. Furthermore, several research questions derived from this exploration are identified along with possible solutions. 

\section{Exploring the state of the art}

\subsection{Related work}
\label{sec:soa}

 Typically, RL methods use the latest observations (from sensors) to improve the behaviour policy of the agent. In other words, the agent interacts with the environment guided by a reward signal. There are two main approaches used to manage how experience (i.e., sensor, reward, and action data) is used in the learning phase to update the policy network's weights; these are called on-policy and off-policy methods. 
 
 In on-policy learning, the agent learns based on rewards achieved by the current policy. This means that the policy to be updated is the same policy that the agent is currently using to determine what actions it takes. In other words, the policy updates with every interaction.

 Alternatively, off-policy RL methods update a policy that is different from the one used to take actions. In this case, the agent interacts with the environment and stores the experience collected in a \textit{replay buffer}, such that the data generated by each new policy $\pi_{k}$ will be kept in the replay buffer until a later time. Then, this replay buffer is used to train the policy $\pi_{k+1}$, meaning experience from $\{\pi_{0}, \pi_{1}... \pi_{k}\}$ is considered when updating $\pi_{k+1}$. 

\textit{Deep Deterministic Policy Gradients (DDPG) \cite{ddpg}} is an off-policy RL method that is composed of two different networks: an actor, that will determine which action the agent performs, and a critic, which will rank how good or bad the action taken is, given a particular environment observation. Firstly, the actor network uses the mean of the values produced by the critic using a maximization objective. Additionally, the critic loss is the mean-squared-error (MSE) of the updated and original $\mathbb{Q}$ values and  the updated $\mathbb{Q}$ value is obtained by the target network.

 

Off-policy reinforcement learning methods use experience replay during their learning phase. Typically, a random mini-batch from this buffer would be sampled and used to update the network, tackling the problem in a similar fashion as supervised learning. There are many different techniques, however, to sample the experience to be used, and it has a direct impact on the learning capabilities of the previously mentioned off-policy RL algorithms. The experience replay technique that achieves better success rate in manipulation tasks is \textit{Hindsight Experience Replay} (HER) \cite{her}.
    
 HER is a replay strategy that can be used with any goal-based off-policy RL algorithm and works very efficiently with sparse rewards. In particular, HER has proven to be efficient when dealing with binary rewards. This replay strategy stores previous experience, and during the sampling phase, a set of sampled transitions have their goal $g$ replaced by $g’$, where $g’$ is an actual goal achieved later in the episode. For example, if we are training a robotic arm to reach a target position A, but instead the agent reaches the point B, in the standard RL framework it would usually get a negative reward or no reward, depending on the reward definition. Before updating the network, HER replaces goal of position A with the actually achieved position B, resulting in positive reward. This information can be used to accelerate the training, and therefore, even when the policy generates random actions, these altered transitions will help the agent learn the optimal policy more quickly.

\subsection{Simulated environment}


Mujoco \cite{mujoco} is a physics engine perfectly suited to robotic manipulation simulation. Most state of the art RL methods are tested on manipulation tasks implemented on Mujoco before moving to more challenging real-world scenarios. Mujoco has a lot of community support and contributions, especially since becoming open-sourced, and there are many environments implemented  by OpenAI Gym \cite{gym} that are based on Mujoco. For all these reasons, the Mujoco simulator was selected to evaluate the RL algorithms explained in Section \ref{sec:soa}.


There are two different robotic systems implemented on OpenAI Gym: the fetch and hand robotics simulators. We first explored HER on the fetch robotic tasks and it was found to be easy to train and it was able to learn fast. We also accelerated training with simple curriculum learning tricks, where we transferred knowledge to both the actor and critic networks from similar tasks using weight initialization (i.e. when learning to push a cube, you implicitly learn how to get to the cube, which can be used for a Pick\&Place task). However, the algorithm was found to struggle more on more complex tasks where more advanced physics were involved, such as the Slide task, where the robot needs to hit a puck with the appropriate force so that it decelerates due to friction and stops at the target position.

While this algorithm was designed to solve this task, it has more difficulty in hand-based environments. In these environments, the goal is to manipulate an object so that it reaches a particular pose and/or position. However, in order to have a successful policy on these manipulation tasks, the algorithm needs to train for more than $38\cdot 10^7$ time steps, which can be infeasible for most researchers. Furthermore, the success on the tasks is not perfect and still leaves room for improvement \cite{mujocopy}.

\subsection{Real Robot Challenge}

Simulators, particularly in the robotics context, are not perfect models of reality. There are multiple factors that make real world scenarios more challenging than a simulated environment. For example, sensors produce noisy data, or robots suffer from vibrations; both things that are not included in simulators. Robotic manipulation in the real world is still, therefore, very challenging. To direct attention towards real-world performance, the robotics community has been very active in organizing competitions like the Real Robot Challenge (RRC) \cite{rrc}.

The RRC is a competition organized by the Max Planck Institute for Intelligent Systems. The 2021 edition consisted of a three-fingered robot that had to carry a cube along specified goal trajectories. The organizers of the challenge provided a simulator of the robot along with remote access to real robots so that participants could run their experiments on a real setup. I decided to participate in the challenge as this was a great opportunity to explore HER on a real robot.

In order to solve the proposed task, a mix of goal-based sparse rewards and dense distance rewards was used. The goal-based sparse reward $r_{xy}$ was used to teach the robot how to move the cube along the surface (xy-plane) (see Eq. \ref{eq:xyonly}), while the role of the dense reward $r_{z}$ was to teach the robot how to lift the cube (see Eq. \ref{eq:z}). To do so, the reward was weighted so that it received a smaller penalty when the cube was above the target location. This reward had to be included due to the difficulties that the standard HER algorithm was having in finding a successful grasping strategy. Both rewards were defined as follows:

\begin{equation}
\label{eq:xyonly}
    r_{xy} =  \begin{cases} 
      0 & \textrm{if} \quad \left\| xy_{cube} - xy_{goal} \right\|\leq 2 cm \\
     -1 & \textrm{otherwise}
   \end{cases}
\end{equation}

\begin{equation}
\label{eq:z}
  r_{z} =  \begin{cases} 
      - 20|\, z_{cube} - z_{goal}|  & \textrm{if} \quad z_{cube} <  z_{goal} \\
      \\
      -10|\, z_{cube} - z_{goal}|  & \textrm{if } \quad z_{cube} > z_{goal}
   \end{cases}
\end{equation}

where ${xy}_{cube}$ are the x-y coordinates of the actual x-y coordinates of the cube, and $xy_{goal}$ are the x-y coordinates of the desired goal, and $z_{cube}$ and $z_{goal}$ are the z-coordinates of the cube and goal, respectively.

The policy was trained applying HER only to $r_{xy}$ and using domain randomization \cite{domain_r} to mimic the noise that the real robot and/or environment might add. This same policy was directly transferred to the real robot without any kind of domain adaptation and was able to successfully solve the task. Our submission outperformed all competitors' submissions and won this phase of the challenge.

\section{Addressing research questions}

Several research questions emerged after experimentation with state of the art techniques. Even though model-based off-policy RL algorithms can learn robotic manipulation tasks from scratch, the difficulty inherent in said tasks implies a huge investment in time and computing resources. This gives rise to a first question: \textit{How can a robot learn manipulation skills faster?}


This was not the only limitation observed. State of the art techniques make use of known priors about the object being manipulated (e.g. weight, pose, shape, etc.), which limits the generalizability of the learned policy. While in simulated environments all this information is perfect, real world setups are noisy and sensors do not work as accurately. Furthermore, real world objects have a wide variety of shapes and weights, which makes manipulation impossible for an RL-based policy that has been trained to manipulate basic shapes such as cubes or spheres. Therefore, previous algorithms must be trained in a more diverse environment if the robot is to interact with the real world. 


Advances in deep learning for computer vision over the last decade along with the emergence of new tactile technologies strongly suggests to consider using both vision and touch as plausible solutions for the above-mentioned problem. Humans make use of vision in order to understand the object to manipulate, and then touch to do the actual manipulation. Hence, we ask ourselves: \textit{How can a robot make use of vision and touch to achieve better dexterity in real world scenarios? How can a robot manipulate a wide variety of objects?}

\subsection{Multi-step robotic manipulation}
\label{sec:multi}

One of the main reasons why state of the art techniques struggle to learn successful policies quickly is because manipulation tasks are usually too complex to be learned easily using sparse rewards. However, most manipulation tasks are a composition of several simpler tasks that are easier to learn. For example, teaching a robot how to reach a place in the space and how to pick an object independently is simpler than teaching a robot a pick \& place task, and as a consequence doing the latter would probably take more time.

This is of particular interest for those tasks which require days of simulated experience in order to achieve a decent performance, like the OpenAI Gym's robotic hand environments. There are three different robotic hand environments in OpenAIGym, depending on the object to manipulate (a cube, a ball and a pen), and each of them has four different tasks with very different complexity depending on the target rotation vector. There is a task that requires the agent to apply random rotations only along the $z$ axis, a second task that requires random rotations along the $z$ axis and axis-aligned rotations for the $x$ and $y$ axes, a third task that does it along all axes, and a fourth task that also adds a target position (includes a translation besides the rotation desired).

The performance of HER drastically changes from the first of the tasks explained above to the second: while learning random rotations only along one axis is fast, when adding the other two rotations, even when those are axes-aligned, the simulated experience needed increases substantially, as the time steps needed to achieve a decent performance move from $2\cdot 10^6$ to more than $38\cdot 10^7$. Furthermore, while HER is able to solve the first task perfectly, this does not happen when adding rotations along all axis as success rate achieved, depending on the object that is being manipulated, ranges from 30\% to 90\%.

However, any rotation can be decomposed into three elementary consecutive rotations around specified fixed-axes that are named Davenport angles \cite{davenport}. This problem can also be understood as finding the matrix decomposition of a rotation given a known coordinate reference system. In the context of the robotic hand environment, any cube rotation can be described as a concatenation of three rotations around the $z$, $x$ and $z$ axes, being only necessary to learn how to rotate the cube around two of the three axes.


\begin{figure}[t]
\centerline{\includegraphics[width=0.45\textwidth]{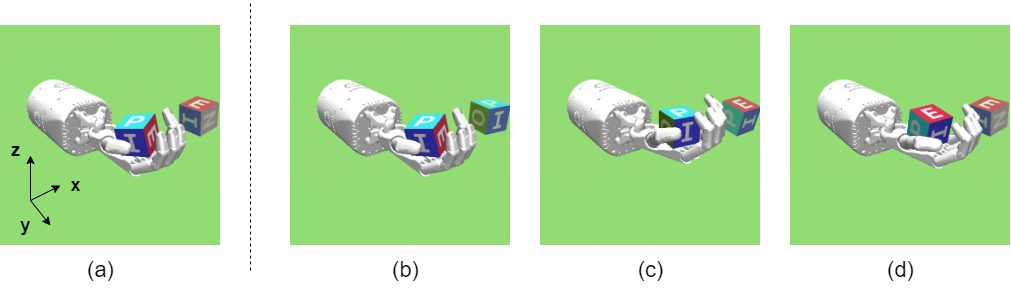}}
\caption{Example of a rotation vector decomposition into three elementary rotations around the $z$, $y$ and again $z$ axes.}
\label{fig}
\end{figure}

\subsection{Exploiting vision}
\label{subsec:vision}


Observations obtained from simulators usually consist of a concatenation of robot data and some kind of information about the object to be manipulated that, in a real setup, is supposed to be captured by different sensors (usually location and pose of the object). This information about the object, however, can also be obtained through vision. Pose estimation and object tracking have been two areas of great interest for the computer vision community, leading to the emergence of a wide variety of techniques capable of solving these problems. Furthermore, the increased interest in robotics in recent years has lead to the creation of datasets containing models of household objects ready to be used in the Mujoco simulator \cite{household}. Considering that images can encode more meaningful information than just a pose descriptor, and the availability of data, it is sensible to utilize vision as a pillar towards improving robotic manipulation. 


There are multiple ways to use vision in a robotic environment. For example, one approach consists of extracting whatever information is necessary from the manipulated object from the image observation (object pose, position, etc), and concatenating this information to the robot data to use as input \cite{obj_det}. Other approaches, instead, encode all this information into a higher dimensional vector by using any state representation learning technique \cite{rl_review}. The use of known world priors can be determined in order to encode information and produce meaningful states (e.g. forcing states that are close to each other in time to be similar) \cite{world_pr}.  


Finding a way to embed meaningful visual information would also answer the second of the research questions proposed in the introduction of this Section. From vision, humans can not only identify what object they want to manipulate, the location and its pose, but they also can make assumptions on how much the object weighs and how much force will be needed to make the grasp. Therefore, this encoding technique should be able to model information of this kind. It is therefore obvious that robots need vision if we want them to be useful in the real world.

\subsection{Exploiting Touch} 
\label{subsec:touch}


Touch is the most important sense during the manipulation phase. Humans use touch in order to grasp the object, detect when it's slipping because they're not applying enough force, etc. Furthermore, there is no real way for a robot to have full certainty it is touching an object without tactile sensing. There are many different tactile sensors, and in recent years there have been a focus on creating simulators that model their behaviour. Therefore, the use of touch in robotic manipulation seems as obvious as the use of vision.


Some of the information useful for manipulation that have been mentioned in Section \ref{subsec:vision} is impossible to obtain without tactile sensing. For example, there is no way to predict how much an object will weight or how much force the robot will need to apply to lift an object without touch. Furthermore, touch can also be used to refine the observation obtained from vision, as a robot can touch an object before manipulating it and obtain additional information about it. 


Usages of touch data in robotic manipulation may vary depending on the type of sensor \cite{digit, feel}. However, touch data is usually used as input data concatenated along robot data, or as an auxiliary source of information (e.g. a successful grasp can be detected through tactile information). There is still a lot to explore on this field as it is a fairly new research field. 

\section{Discussion}
This paper has explored and highlighted the limitations that current deep reinforcement learning techniques have, particularly when solving complex manipulation tasks. Firstly, a mix of goal-based and dense rewards has been proposed in order to solve a manipulation task on a real robot. The addition of the dense reward was necessary because HER was not able to find a policy able to grasp the cube efficiently.

Furthermore, several research questions have been exposed, along with several potential solutions. Current state of the art techniques have a slow learning pace due to the amount of simulated data they need to achieve convergence. A goal-shaping based technique using Davenport angles has been proposed in order to accelerate learning. Additionally, the need of vision and touch has also been exposed, along with several research lines that these could lead to.

\end{document}